\documentclass[conference]{IEEEtran}
\usepackage{cite}
\usepackage{amsmath,amssymb}
\usepackage{graphicx}
\usepackage{booktabs}
\usepackage{array}
\usepackage{multirow}
\usepackage{url}
\usepackage[T1]{fontenc}
\usepackage{textcomp}

\begin{document}

\title{%
  {\LARGE When Does Small Data Work?\par}
  \vspace{0.45em}
  {\normalsize\itshape Accuracy and Efficiency Trade-offs Between Tabular Foundation Models\\
  and Conventional Methods for Crowd-State Classification at Hajj and Umrah\par}

}

\author{
\IEEEauthorblockN{AlJawharh S. AlOtaibi\textsuperscript{*}, Mohamed Eltahir\textsuperscript{*}, Jude AlSubaie*}
\IEEEauthorblockA{
AlJawharh@daldata.ai, M.eltayeb@daldata.ai, Jude@daldata.ai  \\
Riyadh, Saudi Arabia \\
}
\thanks{\textsuperscript{*}These authors contributed equally to this work.}
}

\maketitle

\begin{abstract}
Learning from few labeled examples is a central challenge in tabular machine learning, and it becomes the binding constraint in domains where labeling is costly, such as crowd monitoring during Hajj and Umrah. Tabular foundation models, which predict from only a handful of examples without task-specific training, were recently introduced to address this very-few-label regime. In this study we test them on crowd-state classification to assess how much they help when labels are scarce, and we compare them against standard machine learning methods to characterize the accuracy and efficiency trade-offs between the two approaches. Using three real datasets we evaluate different Machine Learning models, in untuned and tuned forms, against three foundation models. Results show that no single family is best everywhere. The right choice depends on the label budget. When labels are very few, foundation models lead. As labels grow, tuned conventional models catch up and significantly surpass the foundation models on the more structural geometry target. Efficiency separates them further where tuned Machine Learning models incur a large tuning cost that foundation models avoid, although foundation models reprocess their context at every prediction. We summarize these results as a practical map of which approach to prefer under a given label budget and computational budget.
\end{abstract}


\section{Introduction}
Learning from few labeled examples is a fundamental challenge in tabular machine learning. In many high-stakes domains, labels are scarce, costly, or slow to obtain, and a model must perform well from only a handful of examples. Crowd safety during Hajj and Umrah is a stark instance. Each year millions of pilgrims move through a small set of recurring high-risk regimes, including counterflow during Tawaf around the Kaaba, severe bottlenecks at the Jamarat, and dense throughput corridors through Mina and the Mashaer, where local density and flow can cross a critical threshold within moments and the margin between a safe crowd and a dangerous one is narrow. Early, reliable classification of crowd state is therefore central to safety and pilgrim wellbeing. Yet labeled crowd data from the holy sites is scarce, slow to annotate, and rarely available at scale before a season begins, so what matters in practice is not accuracy given abundant labels but accuracy given almost none.

Tabular foundation models have recently been proposed for exactly this regime. Rather than being trained on each dataset, they are pretrained once and adapt to a new task in context, predicting from a small labeled set supplied at inference without task-specific optimization. This in-context ability makes them attractive when labels are few, and a natural candidate for crowd-state classification from tabular features derived from pedestrian trajectories and pilgrim-worn sensors. This route is distinct from the image and video pipelines that dominate crowd intelligence at the holy sites.

Their promise, however, has not been tested against the methods practitioners actually use, under realistic conditions. Strong conventional models, and gradient boosting in particular, remain the default for tabular data and are highly competitive once tuned. It is unclear whether foundation models retain an advantage when conventional baselines are properly tuned, how that advantage changes as the label budget grows, and whether any accuracy gain justifies its computational cost. These questions are decisive for deployment, yet they are seldom answered together.

In this work we study them directly. Using three real crowd datasets spanning controlled crowds, a dense field gathering, and pilgrim-worn sensors, we compare tabular foundation models against a broad set of conventional methods, including tuned gradient boosting, across label budgets ranging from very few to many, under a rigorous, leakage-controlled protocol. We evaluate not only accuracy but also label-efficiency and computational cost, so that the comparison reflects what a deployment actually pays. The result is not a single winner but a map. The preferable family depends on the label budget and the task: when labels are very few, foundation models give the strongest accuracy with no tuning; as labels grow, tuned conventional models close the gap and, on the more structural target, surpass them, while the two families differ sharply in cost.

\noindent\textbf{Contributions.}
\begin{enumerate}
\item A study comparing tabular foundation models against a broad range of established tabular methods, including tuned modern gradient boosting, for crowd-state classification from movement and wearable signals at Hajj and Umrah.
\item A joint evaluation of accuracy, label-efficiency, and computational cost across model families under a rigorous protocol with group-aware splits, permutation-based signal testing, and paired tests with bootstrap confidence intervals.
\item A per-family map of which model type to prefer under which operational conditions, including label budget and inference constraints, intended as practical guidance for label-scarce deployments.
\end{enumerate}

\noindent\textbf{Scope of novelty.} Tabular data mining and wearable sensing have been applied to Hajj before \cite{alshaery2024,fahad2025}. What we add is a systematic, label-efficient comparison of modern tabular foundation models against properly tuned conventional methods for Hajj and Umrah crowd safety, evaluated under scarcity and leakage controls and reported as a conditional, operator-facing map rather than a single verdict.

\section{Background and Related Work}
\subsection{Crowd Safety at Hajj and Umrah}
Crowd-dynamics research on bottleneck and counterflow phenomena \cite{johansson2008}, together with Hajj-specific work on density estimation and crowd management \cite{fahad2025,shah2024,halboob2024}, establishes local density and flow regime as key determinants of crowd safety at the holy sites. Much of this prior crowd-intelligence work relies on image- and video-based analysis \cite{shah2024,halboob2024}. In contrast, this study pursues a tabular, sensor- and trajectory-based route, and frames the same safety variables as a few-label classification problem an operator could run on live feeds, shifting the emphasis from offline analysis to label-efficient, deployable monitoring.

\subsection{Sensing Crowd State from Trajectories and Wearables}
Crowd state can be sensed at two complementary scales. Pedestrian trajectories yield density and flow-geometry features that describe the crowd as a whole \cite{johansson2008,dufour2025}, while pilgrim-worn sensors yield physiological features such as fatigue that describe the individual pilgrim \cite{alshaery2024}. We use both modalities so the safety picture spans the crowd and the person within it.

\subsection{Tabular Foundation Models and In-Context Learning}
Tabular foundation models depart from the conventional train-per-dataset paradigm. Rather than fitting parameters to a single dataset, they are pretrained once on a large collection of synthetic datasets and then predict on new datasets without any task-specific gradient updates \cite{hollmann2025,muller2022}. TabPFN, one of the most prominent models in this class, is a Transformer pretrained on millions of synthetic datasets generated from structural causal model priors, where each synthetic dataset imitates how features relate to a target in real data \cite{hollmann2025}. At inference, the labeled training set and the query rows are passed together as a single input, and the network returns a predictive distribution in one forward pass. This mechanism, in-context learning, is the same principle underlying large language models, where adaptation to a new task occurs through the input context while the network weights remain fixed \cite{hollmann2025,muller2022}.

Architecturally, TabPFN represents each table cell individually and applies a two-way attention scheme, attending across features within a row and across samples within a feature, which makes the model invariant to the ordering of both rows and columns \cite{hollmann2025}. Because the training set is consumed as context rather than learned from, early models in this class were most effective on small-to-medium datasets, with the original model targeting up to roughly 10{,}000 training examples \cite{hollmann2025}; later releases substantially raised this limit \cite{grinsztajn2026}. This property is the reason these models are well matched to the present setting, where labeled crowd-safety data is scarce. More recent models extend the paradigm. TabICL scales in-context learning to larger tables \cite{qu2025,qu2026}, and LimiX models the joint distribution over features and missingness through unified row-and-column modeling \cite{zhang2025}. Gradient boosting remains the principal point of comparison in the tabular foundation model literature, and omitting a well-tuned gradient boosting baseline would understate the strength of conventional methods; we therefore compare against XGBoost \cite{chen2016}, LightGBM \cite{ke2017}, and CatBoost \cite{prokhorenkova2018} alongside the classical baselines established for Hajj crowd-density classification \cite{fahad2025}.

\subsection{Reliable Evaluation with Scarce Labels}
Two evaluation failures are particularly consequential in a safety-critical, low-label setting. The first is data leakage, the introduction of information about the target that would not legitimately be available at prediction time, whether through label-equivalent features or improper data splits. Leakage is a well-documented cause of inflated performance estimates and has been described as one of the most common data-mining mistakes \cite{kaufman2012}. The second is mistaking chance structure for genuine signal, a risk that grows when labels are scarce. Permutation testing addresses this directly by repeatedly retraining on randomly shuffled labels to estimate a null performance distribution; measured performance is considered meaningful only if it exceeds that null \cite{ojala2010}.

\section{Crowd-State Targets and Data}
We classify three safety-relevant crowd-state targets, grounded in three real datasets (Table~\ref{tab:targets}). The experimental axis is the label budget. Each task is evaluated at 16, 64, 256, 1024, and full labels, tracing performance from the scarce-label regime that binds at the holy sites up to the data-rich limit. The three datasets form a deliberate difficulty gradient: clean controlled crowds, a noisy field gathering, and real-Hajj wearable physiology, so the study can report where label-efficient classification holds and where it does not, rather than asserting a single verdict.

\begin{table}[t]
\centering
\caption{Crowd-state targets and their operational meaning.}
\label{tab:targets}
\small
\begin{tabular}{p{1.6cm}p{3.2cm}p{2.0cm}}
\toprule
Safety target & What it tells an operator & Primary data \\
\midrule
Density regime (low/med/high) & Proximity to critical, crush-prone density & J\"ulich, Lyon trajectories \\
Flow geometry & Counterflow / bottleneck / corridor forming & J\"ulich trajectories \\
Pilgrim fatigue & Wellbeing / early-distress signal & Wearable (real Hajj) \\
\bottomrule
\end{tabular}
\end{table}

\begin{figure}[t]
\centering
\includegraphics[width=\linewidth]{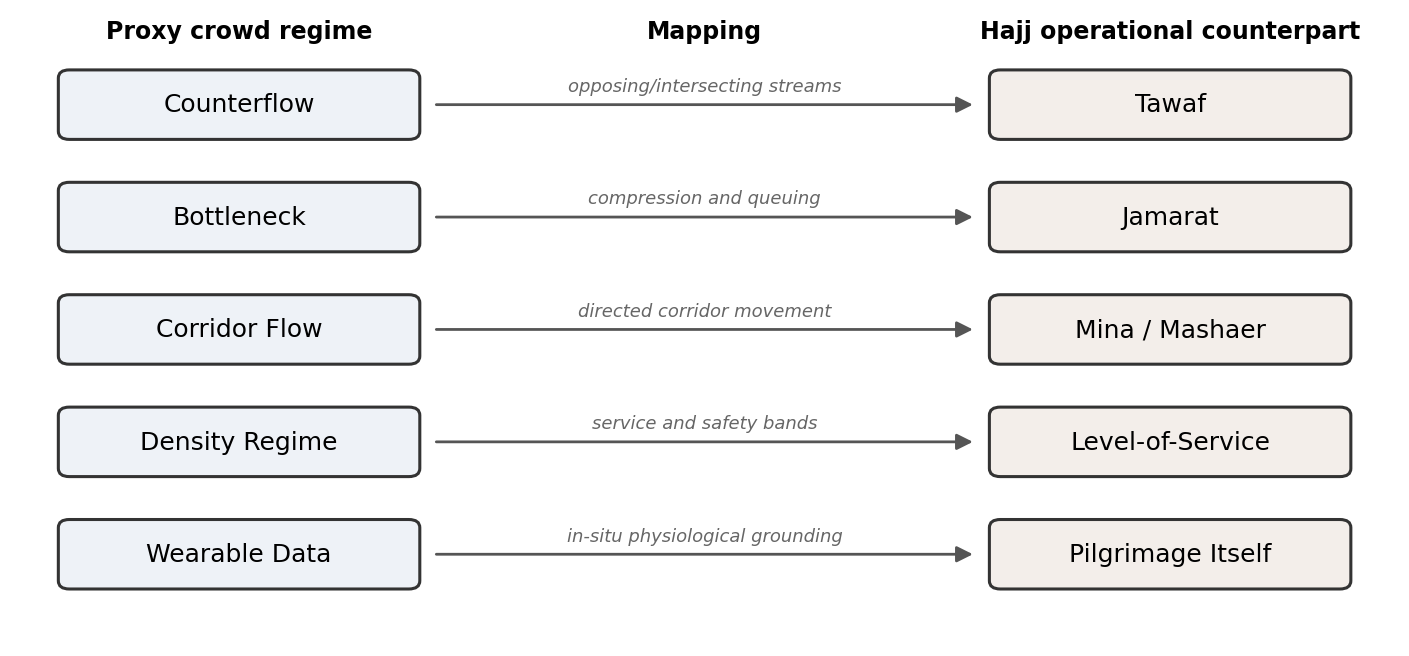}
\caption{Relationship between the evaluated datasets and their Hajj counterparts. Controlled and field datasets act as proxies for recurring Hajj crowd regimes, while the wearable dataset is collected during the pilgrimage itself.}
\label{fig:mapping}
\end{figure}

Each dataset is first reduced to at most 300{,}000 rows, which binds only on the 16.1M-row wearable recording. Within each task, the rows entering any model are further stratified-capped to at most 30{,}000 rows to keep training and evaluation tractable, with stratification preserving each target's class balance so that differences reflect the methods and the budget rather than uneven dataset sizes. Each crowd regime has a direct Hajj counterpart (Table~\ref{tab:datasets}, Fig.~\ref{fig:mapping}). Counterflow maps to Tawaf, bottleneck to Jamarat, corridor to Mina and the Mashaer, and density regime to Level-of-Service. The controlled and field crowds serve as proxies for these Hajj regimes, while the wearable data is recorded at the pilgrimage itself.

\begin{table}[t]
\centering
\caption{Datasets spanning clean, noisy, and real-Hajj conditions (The Rows column is the available feature pool).}
\label{tab:datasets}
\small
\begin{tabular}{p{1.5cm}p{1.4cm}p{2.0cm}r}
\toprule
Dataset & Type & Composition & Rows \\
\midrule
J\"ulich & Real, controlled (lab) & 6 experiments (uni/bi corridor, bottleneck) & 125{,}873 \\
Lyon (Festival of Lights) & Real, field gathering & 6 scenes, $\sim$4{,}800 tracks, 12--30 fps & 200{,}712 \\
Wearable (Al-Shaery) & Real, recorded at Hajj & 64 pilgrims, 17 sensors (16.1M raw) & 300{,}000 (capped) \\
\bottomrule
\end{tabular}
\end{table}

\section{Experimental Design}
\subsection{Classification Tasks and Feature Design}
We cast each safety target as a supervised classification task with a feature set chosen to keep the prediction honest. Density regime is predicted from kinematic features alone (speed and heading alignment); every density-derived quantity is withheld so the model cannot read its own label back out of the inputs, leaving a genuine crush-risk proxy. Flow geometry (counterflow, bottleneck, or corridor) is learned from trajectory features and captures how a regime forms before it turns critical. Pilgrim fatigue is predicted from wearable physiology and stands in for individual wellbeing and early distress.

\subsection{Label Provenance}
A central design choice is that no label is produced by hand. Flow geometry and fatigue come directly from the source datasets, the experiment type and the recorded tiredness field respectively. Density regime is binned deterministically from measured local density and then predicted from kinematics alone, so the columns that define the label never enter the feature set and circularity is removed by construction. The label budgets therefore count how many pre-existing labels the model is shown during in-context learning, not any annotation effort of ours.

\subsection{Evaluation Protocol}
Data splits are group-aware, grouping examples by short time window so that no time window or pilgrim appears in both training and test. This yields several hundred independent groups for the datasets. Each configuration is evaluated over twenty repeated group-disjoint splits, and the same splits are shared across all models so that comparisons are paired. The held-out evaluation set is capped by stratified sampling so that every class remains present. A permutation signal test \cite{ojala2010} screens each task, confirming that measured performance exceeds what shuffled labels would produce. Features are restricted to numeric columns with all label-source fields excluded.

\subsection{Classifiers Under Comparison}
The comparison covers conventional baselines (a majority-class reference, a linear model, nearest neighbors, a support vector machine, a decision tree, and a random forest), three gradient boosting methods, and three tabular foundation models (Table~\ref{tab:classifiers}). Each conventional and gradient boosting method is evaluated in both an untuned default form and a tuned form. Tuning uses a group-aware inner cross-validation carried out within the available label budget, so no tuning advantage leaks across groups or consumes more data than the model is given. The foundation models run at their defaults with no tuning, so any foundation-model advantage is obtained without task-specific tuning. We additionally verify that the conclusions are robust to the preprocessing choice (Section~\ref{sec:robust}).

\begin{table}[t]
\centering
\caption{Classifiers compared.}
\label{tab:classifiers}
\small
\begin{tabular}{p{1.9cm}p{1.5cm}p{2.6cm}}
\toprule
Model & Family & Mechanism \\
\midrule
TabPFN-3 \cite{hollmann2025,grinsztajn2026} & Foundation & PFN, two-way attention, in-context \\
TabICL-v2 \cite{qu2025,qu2026} & Foundation & PFN, scalable to large tables \\
LimiX-2M \cite{zhang2025} & Foundation & Row + column embeddings \\
Random Forest \cite{fahad2025} & Classical & Bagged decision trees \\
XGBoost \cite{chen2016} & Gradient boosting & Regularized, level-wise \\
LightGBM \cite{ke2017} & Gradient boosting & Leaf-wise, histogram-based \\
CatBoost \cite{prokhorenkova2018} & Gradient boosting & Ordered boosting \\
LogReg / KNN / SVM / DT \cite{fahad2025} & Classical & Linear / instance / margin / tree \\
Dummy & Classical & Majority class \\
\bottomrule
\end{tabular}
\end{table}

\subsection{Scoring and Significance Testing}
Performance is reported as macro-F1, averaged over twenty repeated group-disjoint splits, with a majority-class baseline to confirm models exceed chance. Each method is summarized three ways. We report accuracy at each label budget, label-efficiency as the area under the low-label performance curve, and computational cost as wall-clock time to fit and predict. Differences from the strongest conventional model use the paired Wilcoxon signed-rank test with effect sizes and bootstrap confidence intervals, and we apply both a conservative correction across all tests and a focused correction for the primary low-label comparison.

\section{Results}
\subsection{Crowd-State Signal is Genuine}
Every task passes the permutation signal test with a valid group split (Table~\ref{tab:signal}). The density and geometry carry strong real-versus-null separation while wearable fatigue is at risk, so it will be discarded for the rest of the work. The p-value floor of 0.0099 is set by the 100-permutation null and is identical across rows by construction; the discriminating quantity is the gap between the real and null macro-F1, not the p-value.

\begin{table}[t]
\centering
\caption{Permutation signal test, real versus shuffled-label null. The p-value floor (0.0099) is set by the permutation count.}
\label{tab:signal}
\small
\begin{tabular}{lrrr}
\toprule
Target / dataset & Real F1 & Null mean & p \\
\midrule
Density / J\"ulich & 0.6529 & 0.3006 & 0.0099 \\
Flow geometry / J\"ulich & 0.7630 & 0.1781 & 0.0099 \\
Density / Lyon & 0.4495 & 0.3063 & 0.0099 \\
Fatigue / wearable & 0.1331 & 0.1313 & 0.0099 \\
\bottomrule
\end{tabular}
\end{table}

\subsection{Which Family Leads, by Budget and Task}
\begin{figure}[t]
\centering
\includegraphics[width=\linewidth]{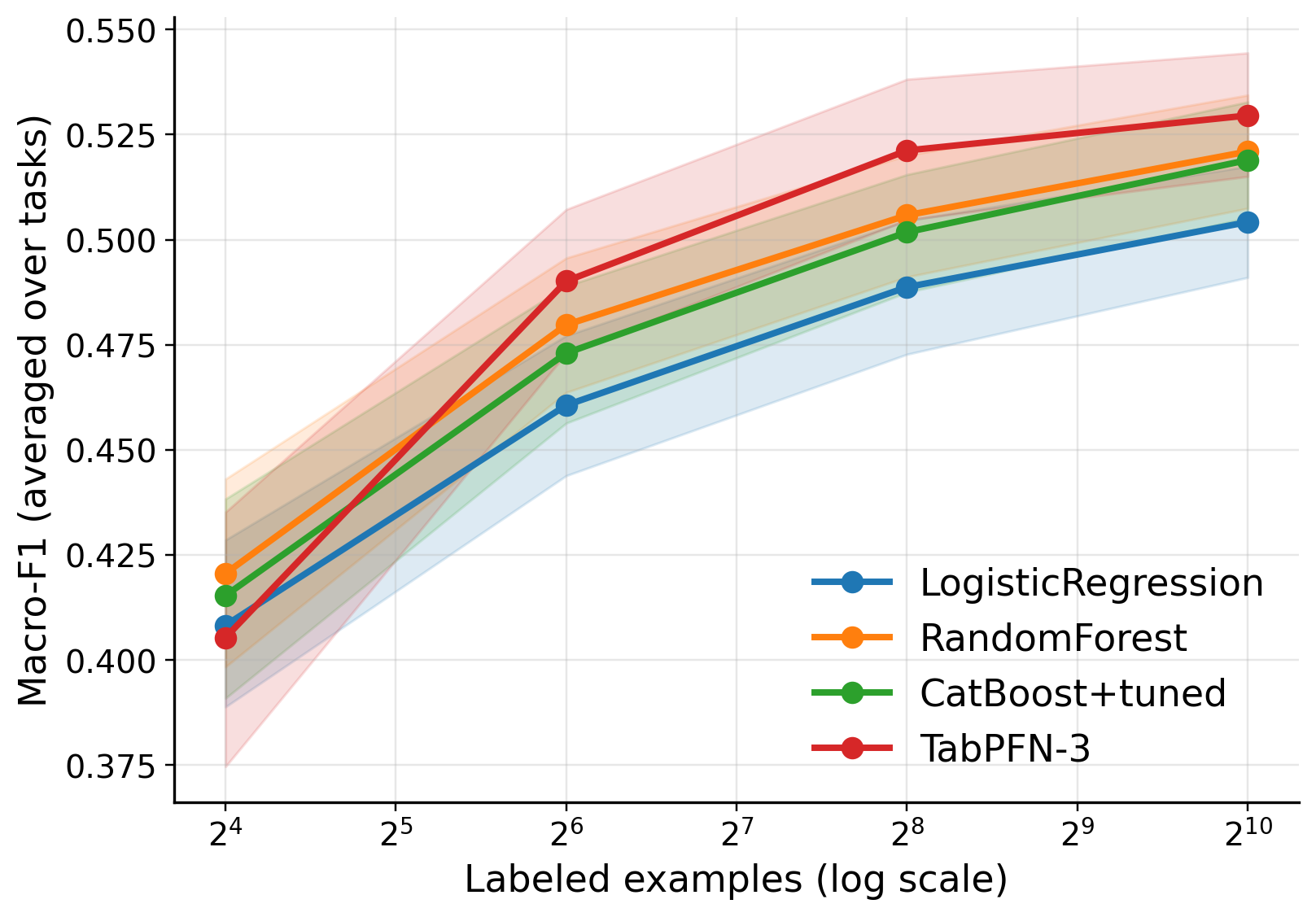}
\caption{Macro-F1 as a function of labeled examples, averaged across tasks. The foundation model leads in the low-label region and conventional methods converge as labels increase. Values are averaged over tasks of differing difficulty, so the ordering and shape, rather than the level, carry the message.}
\label{fig:curves}
\end{figure}

At the smallest budgets a foundation model tends to lead; at the largest budget tuned conventional models lead; and the size of any advantage depends on the task (Table~\ref{tab:regime}). The one strong, supported low-label result is on the density target, where the foundation model is the most label-efficient method and significantly outperforms even tuned gradient boosting at low budgets (Table~\ref{tab:focused}). On the geometry target, tuned tree-based models are at least as strong at every budget. Fig.~\ref{fig:curves} shows the learning curves averaged across tasks.

\begin{table}[t]
\centering
\caption{Regime overview of which family leads at the smallest and largest budget, and which leads on integrated label-efficiency (AULC). ``Significant on AULC'' marks a statistically supported foundation-model advantage in integrated low-label accuracy.}
\label{tab:regime}
\footnotesize
\setlength{\tabcolsep}{3pt}
\begin{tabular}{p{1.3cm}p{1.0cm}p{1.5cm}p{1.4cm}p{1.3cm}}
\toprule
Dataset & Task & Fewest labels & Full data & Most label-eff. \\
\midrule
Controlled & Density & Foundation & Linear & Foundation (sig.\ on AULC) \\
Controlled & Geometry & Grad.\ boost & Grad.\ boost (tuned) & Foundation (n.s.) \\
Field & Density & Linear & Tree ens.\ (tuned) & Foundation (n.s.) \\
\bottomrule
\end{tabular}
\end{table}

\begin{table}[t]
\centering
\caption{Focused, pre-registered comparison at low budgets, TabPFN-3 versus tuned gradient boosting (CatBoost+tuned). Positive favors the foundation model. The density advantage is significant after Holm correction at every low budget; geometry is comparable.}
\label{tab:focused}
\footnotesize
\setlength{\tabcolsep}{4pt}
\begin{tabular}{llrrrrc}
\toprule
Task & Lab. & Found. & Tn.\,GB & Diff. & Adj.\ $p$ & Sig. \\
\midrule
Density & 16 & 0.588 & 0.556 & $+0.032$ & 0.020 & yes \\
Density & 64 & 0.648 & 0.615 & $+0.035$ & 0.001 & yes \\
Density & 256 & 0.665 & 0.637 & $+0.024$ & $<0.001$ & yes \\
Geometry & 16 & 0.555 & 0.583 & $-0.012$ & 0.234 & no \\
Geometry & 64 & 0.671 & 0.661 & $+0.005$ & 0.312 & no \\
Geometry & 256 & 0.719 & 0.708 & $+0.009$ & 0.065 & no \\
\bottomrule
\end{tabular}
\end{table}

\subsection{Label-Efficiency, and a Non-Monolithic Family}
Integrated across the low-label range, the strongest foundation model is the most label-efficient method on the clean density target, and this advantage is significant even against the best tuned conventional model (Table~\ref{tab:aulc}). On every other task the differences are not significant. Crucially, the foundation family is not interchangeable. Reporting all three models reveals a significant \emph{loss}. On geometry, LimiX-2M trails the best tuned baseline by 0.013 in integrated accuracy after correction. Claims should therefore be made about specific models, not the category as a whole.

\begin{table}[t]
\centering
\caption{Integrated low-label accuracy (AULC) of each foundation model against the best tuned conventional model on the same task. Positive favors the foundation model. Significance is after Holm correction.}
\label{tab:aulc}
\footnotesize
\setlength{\tabcolsep}{4pt}
\begin{tabular}{llrrrc}
\toprule
Dataset & Task & Found.\ model & AULC & Diff. & Sig. \\
\midrule
Controlled & Density & TabPFN-3 & 0.648 & $+0.010$ & yes \\
Controlled & Density & TabICL-v2 & 0.636 & $+0.002$ & no \\
Controlled & Density & LimiX-2M & 0.634 & $-0.001$ & no \\
Controlled & Geometry & TabPFN-3 & 0.681 & $+0.008$ & no \\
Controlled & Geometry & LimiX-2M & 0.658 & $-0.013$ & loss \\
Field & Density & TabPFN-3 & 0.466 & $+0.001$ & no \\
\bottomrule
\end{tabular}
\end{table}

\subsection{Full-Budget Comparison}
At full budget the best conventional model is at least as strong as the foundation models on every task, and significantly stronger on several. The clearest case is geometry, where all three foundation models lose to tuned gradient boosting after correction, by 0.096 (LimiX-2M), 0.033 (TabICL-v2), and 0.022 (TabPFN-3). LimiX-2M also loses significantly at full budget on density and on Lyon (Table~\ref{tab:fullbudget}). The value of foundation models is therefore concentrated in the low-label regime, not at full data, and significance runs in both directions, a foundation win at low-label density and foundation losses at full-budget geometry.

\begin{table}[t]
\centering
\caption{Full-budget paired comparison of each foundation model against the strongest conventional model. Negative favors the conventional model. Significance is after Holm correction.}
\label{tab:fullbudget}
\footnotesize
\setlength{\tabcolsep}{4pt}
\begin{tabular}{llrrc}
\toprule
Task & Found.\ model & Best conv. & Diff. & Sig. \\
\midrule
Geometry & LimiX-2M & CatBoost+tuned & $-0.096$ & loss \\
Geometry & TabICL-v2 & CatBoost+tuned & $-0.033$ & loss \\
Geometry & TabPFN-3 & CatBoost+tuned & $-0.022$ & loss \\
Density & LimiX-2M & LogisticRegression & $-0.009$ & loss \\
Lyon dens. & LimiX-2M & RandomForest+tuned & $-0.022$ & loss \\
\bottomrule
\end{tabular}
\end{table}

\subsection{Effect of Tuning}
Tuning conventional models helps most at the largest label budget and rarely at the smallest, where there is too little data for the tuning search to be reliable. At low budgets, foundation models require no tuning yet match or exceed tuned conventional models on the density target. This supports the operational point that foundation models are attractive precisely when labels are few and tuning is not feasible.

\subsection{Computational Cost}
\begin{figure}[t]
\centering
\includegraphics[width=\linewidth]{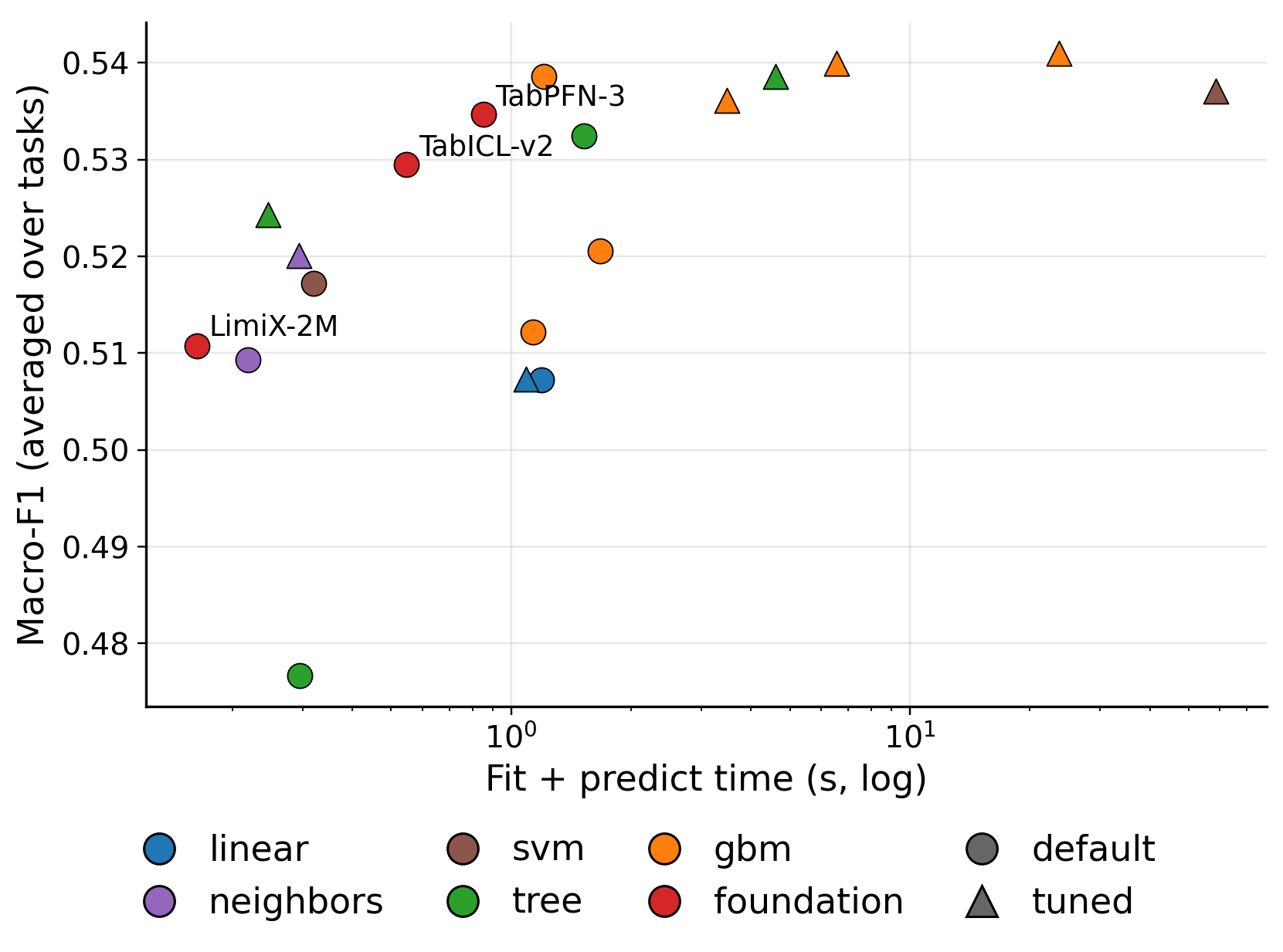}
\caption{Cost against accuracy at the largest budget, averaged across tasks, colored by family (circle is default, triangle is tuned). Points toward the upper left are more accurate and cheaper. Foundation models sit in the accurate and inexpensive region, while tuned gradient boosting and support vectors are far more costly for comparable accuracy.}
\label{fig:speed}
\end{figure}

Conventional models are trained once and are then cheap to query, while foundation models avoid training but reprocess their full context at prediction time, so their cost grows with deployment scale. Tuning a conventional model adds a one-time search cost that can be substantial. At the scale studied, the foundation models are competitive in cost with untuned gradient boosting and far cheaper than tuned gradient boosting (Table~\ref{tab:latency}, Fig.~\ref{fig:speed}). LimiX-2M predicts in 0.14\,s at accuracy comparable to gradient boosting, while tuned gradient boosting and tuned support vectors cost 15.0\,s and 56.1\,s respectively for no accuracy gain on this target.

\begin{table}[t]
\centering
\caption{Representative fit-and-predict time and accuracy at the largest budget on the density target. Times for tuned methods include the one-time tuning search. Foundation models require no tuning.}
\label{tab:latency}
\small
\begin{tabular}{llrr}
\toprule
Method & Family & Time (s) & Acc. \\
\midrule
Nearest neighbors & conventional & 0.09 & 0.646 \\
LimiX-2M & foundation & 0.14 & 0.674 \\
Linear model & conventional & 0.33 & 0.683 \\
TabICL-v2 & foundation & 0.48 & 0.677 \\
TabPFN-3 & foundation & 1.00 & 0.678 \\
Gradient boosting & gradient boosting & 1.08 & 0.674 \\
Random forest & conventional & 1.49 & 0.673 \\
Gradient boosting (tuned) & gradient boosting & 14.99 & 0.681 \\
Support vector (tuned) & conventional & 56.06 & 0.670 \\
\bottomrule
\end{tabular}
\end{table}

\subsection{What Each Family is Good For}
Table~\ref{tab:profile} summarizes accuracy at the smallest and largest budget, integrated label-efficiency, typical cost, and tuning gain for each family on the two clean tasks. Foundation models lead on label-efficiency on the density target and are competitive in cost with untuned conventional models. Tuning helps conventional models most at the largest budget; at the smallest budget there is too little data for tuning to be reliable, which is precisely where foundation models are most useful.

\begin{table}[t]
\centering
\caption{Per-family summary on the two clean tasks. For each family the best representative is shown. Foundation models require no tuning.}
\label{tab:profile}
\scriptsize
\setlength{\tabcolsep}{3pt}
\begin{tabular}{llrrrrr}
\toprule
Task & Family & Acc.\ few & Acc.\ full & L-eff. & Cost (s) & Tun.\ gain \\
\midrule
Density & Linear & 0.562 & 0.683 & 0.633 & 0.38 & 0.000 \\
Density & Neighbors & 0.503 & 0.667 & 0.592 & 0.16 & 0.019 \\
Density & SVM & 0.558 & 0.670 & 0.616 & 28.16 & 0.015 \\
Density & Tree ens. & 0.574 & 0.682 & 0.634 & 0.88 & 0.049 \\
Density & Grad.\ boost & 0.560 & 0.683 & 0.619 & 2.02 & 0.020 \\
Density & Foundation & 0.588 & 0.678 & 0.648 & 0.48 & n/a \\
\midrule
Geometry & Linear & 0.532 & 0.671 & 0.621 & 0.72 & 0.000 \\
Geometry & Neighbors & 0.479 & 0.723 & 0.587 & 0.21 & $-0.001$ \\
Geometry & SVM & 0.506 & 0.776 & 0.625 & 20.66 & 0.057 \\
Geometry & Tree ens. & 0.579 & 0.778 & 0.676 & 1.05 & 0.018 \\
Geometry & Grad.\ boost & 0.584 & 0.780 & 0.676 & 2.73 & 0.015 \\
Geometry & Foundation & 0.556 & 0.761 & 0.681 & 0.49 & n/a \\
\bottomrule
\end{tabular}
\end{table}

\subsection{Robustness}
\label{sec:robust}
Across three preprocessing variants the change in macro-F1 is small ($|\Delta| \leq 0.012$; e.g., on J\"ulich density, Random Forest $+0.0006$, TabPFN-3 $-0.0043$, TabICL-v2 $-0.0082$ under quantile encoding), and group-aware hyperparameter tuning does not alter the ranking. The conclusions are robust to configuration rather than artifacts of it.

\section{Discussion}
The picture that emerges is a conditional one. Small data does work, but how much it works depends on the task and on the data in front of you, not on the model class by itself.

\noindent\textbf{When foundation models help.} On the clean density target, the foundation model is the most label-efficient method we tested. It reaches macro-F1 0.588 at 16 labels against 0.556 for tuned gradient boosting, and the focused, pre-registered comparison stays significant after correction through 256 labels, where the gap is 0.665 versus 0.637 (Table~\ref{tab:focused}). The advantage is not a single lucky budget. It holds at 16, 64, and 256 labels and in integrated label-efficiency, 0.648 against 0.634 for the best tuned conventional model (Table~\ref{tab:aulc}). What this buys an operator is narrow but real. When labels are scarce and there is no time or expertise to tune, a foundation model at its defaults is a strong place to start on clean sensor-derived tables, with no search and no per-dataset training to set up.

\noindent\textbf{When conventional models are the safer choice.} The advantage does not generalize. On geometry the tuned tree ensembles are at least as strong at every budget, and at 16 labels gradient boosting leads, 0.583 against 0.555. On Lyon, the noisy field gathering, the foundation edge disappears entirely, with integrated label-efficiency essentially tied (0.466 against 0.469) and not significant. At full budget the ordering flips decisively toward conventional methods. On geometry all three foundation models lose to tuned gradient boosting after correction, by 0.022 to 0.096 (Table~\ref{tab:fullbudget}). The value of foundation models is therefore concentrated in one corner of the operating space, the low-label clean-data corner, and shrinks, and on the structural target reverses, as either condition is relaxed.

\noindent\textbf{The family is not monolithic.} Among foundation models, results vary enough that the category label is misleading. TabPFN-3 carries the only significant low-label win, while LimiX-2M is significantly weaker than tuned baselines on geometry at both the smallest and the largest budget. An operator choosing a tool should select a specific model, not a family.

\noindent\textbf{Cost reinforces the recommendation.} Accuracy alone understates the case for foundation models, because the cost profiles differ sharply. At the largest budget the foundation models predict in 0.14 to 1.00\,seconds, in the same range as untuned gradient boosting at 1.08\,seconds and well below tuned gradient boosting at 14.99\,seconds and tuned support vectors at 56.06\,seconds (Table~\ref{tab:latency}). A foundation model buys low-label accuracy without the one-time tuning search that conventional models need to stay competitive, and it does so at default settings. The caveat is scale. Foundation models reprocess their context at prediction time, so for continuous high-throughput monitoring a trained conventional model may regain the cost advantage once enough labels exist.

\noindent\textbf{Practical guidance.} When labels are very few and there is no time or expertise to tune, use a label-efficient foundation model, specifically the strongest member rather than the family by default, as the starting point for crowd-state classification on clean sensor-derived tables. Retain tuned tree ensembles and gradient boosting for noisy field conditions and for any setting where a larger labeled set is available, since they are at least as reliable there and significantly better on the structural geometry target.

\section{Limitations}

\begin{itemize}
\item \textbf{External validity.} The datasets approximate pilgrimage conditions rather than fully capturing them, so the findings are a step toward crowd-safety tooling rather than a fielded system, and their transfer to live operation remains to be established.
\item \textbf{Conditional scope.} The observed advantages are specific to particular tasks and data conditions and do not generalize across all settings. The results should be read as a map of where each family is preferable, not as a universal ranking.
\item \textbf{Statistical versus operational significance.} Performance above chance is not the same as operational usefulness. A deployed tool would need reliable detection of the most dangerous conditions, which the present results do not yet guarantee.
\item \textbf{Cost at scale.} The efficiency comparison reflects the scale studied, and the relative cost of the approaches can shift at larger deployment scale.
\end{itemize}

\section{Conclusion}
We asked when small data works for crowd-state classification at Hajj and Umrah, and the answer is conditional. No single model family is best everywhere. When labels are very few and tuning is not feasible, tabular foundation models are the strongest and simplest choice, reaching useful accuracy at their defaults with no task-specific training. As the labeled set grows or the signal becomes noisier, well-tuned conventional methods, gradient boosting in particular, match or surpass them, most clearly on the more structural task. Efficiency points the same way, since foundation models obtain their low-label accuracy without a tuning search, with the caveat that their cost grows with deployment scale.

\section*{Reproducibility and Data Availability}
All datasets are public and cited by DOI \cite{alshaery2024,dufour2025}; the J\"ulich pedestrian experiments are from the public data archive. Evaluation uses fixed seeds, group-aware splits by time window, a stratified evaluation cap, 100-permutation signal tests, three preprocessing variants, group-aware hyperparameter tuning, paired tests with effect sizes and bootstrap confidence intervals, both a conservative and a focused multiple-comparison correction, conventional and gradient boosting baselines in untuned and tuned forms, and three foundation models. A resumable per-(dataset, task, model) runner stitches slices into the full results, and a notebook with a frozen environment specification and the full per-method results accompanies the paper.

\end{document}